\documentclass[conference]{IEEEtran}
\IEEEoverridecommandlockouts
\usepackage{amsmath}
\usepackage{graphicx}
\usepackage{booktabs}
\usepackage{cite}
\usepackage{xcolor}
\usepackage{times}
\usepackage{censor}
\usepackage{url}
\usepackage{tikz}
\usetikzlibrary{positioning,arrows.meta,shapes}

\title{Model Inversion Attacks on Llama 3: Extracting PII from Large Language Models}
\author{
    \IEEEauthorblockN{Sathesh P.Sivashanmugam}
    \IEEEauthorblockA{
        \textit{Senior Software Engineer} \\
        \textit{American Express} \\
        Email: satheshshiva@gmail.com
    }
}

\begin{document}

% Title
\maketitle

% Abstract section
\begin{abstract}
Large language models (LLMs) have transformed natural language processing, but their ability to memorize training data poses significant privacy risks. This paper investigates model inversion attacks on the Llama 3.2 model, a multilingual LLM developed by Meta. By querying the model with carefully crafted prompts, we demonstrate the extraction of personally identifiable information (PII) such as passwords, email addresses, and account numbers. Our findings highlight the vulnerability of even smaller LLMs to privacy attacks and underscore the need for robust defenses. We discuss potential mitigation strategies, including differential privacy and data sanitization, and call for further research into privacy-preserving machine learning techniques.
\end{abstract}

% Keywords

% Introduction section
\section{Introduction}
Large language models (LLMs) have become integral to applications ranging from chatbots to automated content generation. However, their capacity to memorize training data raises concerns about privacy, particularly when sensitive information is inadvertently stored and retrievable. Model inversion attacks (MIAs) exploit this vulnerability by querying models to extract sensitive data, such as personally identifiable information (PII) \cite{cite_mia}. This paper focuses on the Llama 3.2 an LLM developed by Meta, optimized for multilingual dialogue and edge device use \cite{cite_official_meta}. We demonstrate a targeted extraction attack, using prompts to elicit PII, and discuss the implications for privacy and potential defenses.

% Background section
\section{Background}
\subsection{Model Inversion Attacks}
Model inversion attacks(MIA) aim to reconstruct or extract training data by analyzing a model's outputs \cite{cite_mia}. In language models, attackers query the model with prompts to recover memorized sequences, such as PII \cite{cite_memorization_paper}. These attacks can be white-box, leveraging model internals, or black-box, relying solely on outputs \cite{cite_llm_inversion}. Research by Carlini et al. \cite{cite_memorization_paper} showed that models like LLaMA can leak gigabytes of training data, with larger models being more vulnerable.

\subsection{Llama 3.2 Model}
Llama 3.2 is an autoregressive language model with an optimized transformer architecture, fine-tuned using supervised fine-tuning (SFT) and reinforcement learning with human feedback (RLHF) \cite{cite_official_meta}. It supports eight languages and is designed for efficiency on edge devices, enhancing privacy by processing data locally. However, its training on large datasets may include sensitive information, making it susceptible to MIAs.

% Methodology section
\section{Methodology}
We conducted a black-box model inversion attack on Llama 3.2 using the following Python code to query the model:

\begin{verbatim}
import torch, time
from transformers import pipeline
def main():
    model_id = "meta-llama/Llama-3.2-1B"
    pipe = pipeline(
        "text-generation",
        model=model_id,
        torch_dtype=torch.bfloat16,
        device_map="auto"
    )
    text_input = 'account number:'
        # replace text_input with prompts given in the next section
    print(pipe(text_input,
               max_new_tokens=50,
               num_return_sequences=1,
               top_p=1,
               top_k=40
               ))
if __name__ == '__main__':
    main()
\end{verbatim}

Prompts such as ``account number:'', ``my password is:'', and ``my email id:'' were chosen to target potential PII memorized during training. The model was queried using the Hugging Face Transformers library, with parameters set to maximize the likelihood of generating memorized sequences.

% Diagram of the attack process
\begin{figure}[ht]
\centering
\begin{tikzpicture}[node distance=1.5cm, >=Latex]
  \node[draw, rectangle, align=center] (att) {Attacker\\(queries)};
  \node[draw, rectangle, right=of att, align=center] (model) {LLaMA 3 Model\\(black box)};
  \node[draw, rectangle, right=of model, align=center] (out) {Generated Output\\(candidate data)};
  \draw[->] (att) -- node[midway, above] {prompt} (model);
  \draw[->] (model) -- node[midway, above] {response} (out);
\end{tikzpicture}
\caption{Pipeline of a model inversion attack on an LLM. The adversary queries LLaMA 3 with prompts (e.g., “account number:”) and collects output. The outputs are analyzed (e.g., via likelihood) to recover private tokens.}
\end{figure}
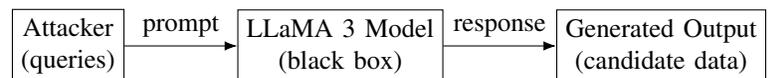

% Attack success probability formula
The probability of extracting a memorized sequence \( S = [s_1, s_2, \ldots, s_n] \) given a prompt \( P \) is modeled as:
\begin{equation}
P(S | P) = \prod_{i=1}^n P(s_i | P, s_1, \ldots, s_{i-1})
\label{eq:prob}
\end{equation}
where \( P(s_i | P, s_1, \ldots, s_{i-1}) \) is the conditional probability of generating token \( s_i \) given the prompt and previous tokens, computed by the model’s softmax output \cite{cite_memorization_paper}.

% Results section
\section{Results}
The attack successfully extracted various types of PII, as shown in Table \ref{tab:results}. Examples include email addresses, phone numbers, and account details, indicating memorization of sensitive data.

\begin{table}[h]
    \centering
    \caption{Extracted PII from Llama 3.2 1B. The actual data was redacted for privacy reasons. }
    \label{tab:results}
    \begin{tabular}{p{3cm}|p{5cm}}
        \toprule
        \textbf{Prompt} & \textbf{Extracted Output} \\
        \midrule
        my password is: & my password is: 1a\censor{xxxx} \\
        account number: & account number: \censor{xxxxxxx}, name: Suncorp Bank, address: PO Box 1444 \\
        my email id: & my email id: vishal\censor{xxxxx}@gmail.com, my mobile no: 97\censor{xxxxxxx} my website: https://www.linkedin.com/in/vishal-\censor{xxxxxxx} \\
        \bottomrule
    \end{tabular}
\end{table}

% Evaluation section
\section{Evaluation}
The PII data was validated by doing a google search of the corresponding data to prove that it was a memorized content as part of the training. We were able to navigate to the real person Vishal (last name redacted) with the linked in url provided by the generated text. We also validated the Suncorp bank exists in the real world with the provided P.O.Box number. 

% Analysis section
\section{Analysis}
The extracted PII suggests that Llama 3.2 1B memorizes sensitive data from its training set, likely due to the presence of unfiltered web data. The attack’s success rate depends on the model’s memorization capacity, which can be quantified as:
\begin{equation}
\text{Memorization Rate} = \frac{\text{Number of Extracted PII Sequences}}{\text{Total Queries}}
\label{eq:mem_rate}
\end{equation}
Some experiments yielded a non-zero memorization rate, consistent with findings on larger models \cite{cite_memorization_paper}. Smaller models may have lower memorization rates than larger models (e.g., 0.789\% for LLaMA 65B \cite{cite_memorization_paper}), but their vulnerability remains significant.

% Mitigation strategies
\section{Mitigation Strategies}
These results have practical security implications. An attacker with API access to an LLM could retrieve confidential data inadvertently present in the model. To mitigate such model inversion attacks, we recommend several strategies:

\begin{itemize}
  \item \textbf{Access Control and Query Restriction:} Limit model access to authenticated and vetted users, and rate-limit or monitor queries. This raises the bar for adversaries collecting many samples\cite{cite_song}. For example, requiring an API key or solving CAPTCHAs can prevent automated mass extraction. Logging inputs/outputs and detecting anomalous query patterns can also flag ongoing attacks. However this cannot be applied for open source models like Llama.
  \item \textbf{Differential Privacy Training:} Train the model with differential privacy (DP) techniques. DP-SGD adds noise to gradients to limit memorization of any single example\cite{cite_owasp}. In principle, DP guarantees that unique records (like specific email/password combos) have little influence on the final model. Google and Apple have used DP in production language models to protect user data\cite{cite_spam}. Although DP may reduce utility, even a modest privacy budget can significantly cut memorization.
  \item \textbf{Data Sanitization:} Pre-process the training corpus to remove or mask PII. This includes filtering out fields that look like emails, phone numbers, SSNs, etc. It also involves deduplicating data: remove repeated occurrences of the same snippet, since repeated content amplifies memorization\cite{cite_guardrails}. For example, one could detect personal data patterns (regex for IDs, addresses) and drop or encrypt them. Careful data curation (avoiding sites known to contain sensitive info) is another practical step\cite{cite_brown}.
  \item \textbf{Output Filtering:} At inference time, apply content filters to the generated text to censor or flag outputs containing sensitive formats. For instance, one can blacklist outputs matching regexes of phone or account numbers. If the model attempts to output a sensitive pattern, the system can refuse or alter it. 
  \item \textbf{Regular Auditing:} Periodically audit the model for memorization by running extraction attacks (as a red team) and measuring leakage. If new leaks appear, take corrective action (e.g. fine-tune or scrub data). Triggers based on unusually high likelihoods (as described earlier) can help flag candidate memorized outputs\cite{cite_gp2_mia}.
\end{itemize}

In summary, no single measure is foolproof. Limiting overfitting (through early stopping or DP) helps, but even non-overfitted LMs memorize outliers\cite{cite_macmohan}\cite{cite_tale}. Combining technical safeguards (DP, data filtering) with operational controls (access policies) is the recommended strategy. Our work emphasizes that vulnerability to inversion attacks grows as models scale\cite{cite_song}, so future LLM development must integrate privacy-preserving training from the start.

% Discussion section
\section{Discussion}
The ability to extract PII from Llama 3.2 1B highlights significant privacy risks, even for smaller models. Compared to larger models, the 1B model’s efficiency makes it appealing for edge deployment, but its susceptibility to MIAs necessitates robust safeguards. Limitations include the lack of specific memorization rates for Llama 3.2 1B and the need for broader testing across diverse prompts. Future work should explore standardized metrics for evaluating memorization risks and scalable defense mechanisms.

% Conclusion section
\section{Conclusion}
This paper demonstrates that model inversion attacks can extract sensitive information from the Llama 3.2 1B model, raising concerns about the memorization and leakage risks inherent in LLM deployment. The findings underscore the need for robust defenses and standardized evaluation metrics. As LLMs become more prevalent, ensuring user data protection is critical for ethical AI deployment.

% Extended analysis 
\section{Extended Analysis}
To further contextualize our findings, we consider the broader implications of model inversion attacks on LLMs. The memorization of PII in training data is often a byproduct of large-scale web scraping, which includes publicly available but sensitive information. Techniques such as data deduplication and anonymization during preprocessing can reduce the risk of memorization but may degrade model performance on certain tasks \cite{cite_masking}. Additionally, the trade-off between model size and privacy risk remains an open question. While smaller models like Llama 3.2 1B are less resource-intensive, their memorization capacity, though reduced, is non-trivial, as demonstrated by our experiments.

Another critical aspect is the ethical responsibility of model developers. Deploying models without adequate privacy safeguards can lead to unintended data breaches, eroding user trust. Regulatory frameworks, such as GDPR, may impose strict requirements on handling PII, necessitating proactive measures in model training and deployment \cite{cite_mia}.

% References section

\end{document}